\documentclass[letterpaper]{article}
\usepackage{graphicx}
\usepackage{spconf}
\usepackage[latin9]{inputenc}
\usepackage{array}
\usepackage{amsmath}
\usepackage{amssymb}
\usepackage{fancyhdr}

\usepackage{subcaption}

\usepackage{nomencl}

\usepackage[english]{babel}

\makeatother

\begin{document}


\title{Microscopy Cell Segmentation via Adversarial Neural Networks }
\name{Assaf Arbelle and Tammy Riklin
  Raviv\sthanks{This study was partially supported by the
    Negev scholarship at Ben-Gurion University (A.A.); The Kreitman
    School of Advanced Graduate Studies (A.A) ; The Israel Ministry of Science, Technology and Space
 (MOST 63551 T.R.R.)}}

\address{Department of Electrical and Computer Engineering \\ Zlotowski Center for Neuroscience\\ Ben-Gurion University of
the Negev, Beer-Sheva, Israel}
\maketitle

\begin{abstract}
We present a novel method for cell segmentation in microscopy images which is inspired by the Generative Adversarial Neural Network (GAN) approach. Our framework is built on a pair of two competitive artificial neural networks, with a unique architecture, termed Rib Cage, which are trained simultaneously and together define a min-max game resulting in an accurate segmentation of a given image. Our approach has two main strengths, similar to the GAN, the method does not require a formulation of a loss function for the optimization process. This allows training on a limited amount of annotated data in a  weakly supervised manner. Promising segmentation results on real fluorescent microscopy data are presented. The code is freely available at: https://github.com/arbellea/DeepCellSeg.git 
\end{abstract}
\section{Introduction}

\label{sec:Introduction}
Live cell microscopy imaging is a key component in the biological
research process. However, without the proper analysis tools, the
raw images are a diamond in the rough. One must obtain the segmentation
of the raw images defining the individual cells prior to calculation
of the cells' properties. Manual segmentation is infeasible due to
the large quantity of images and cells per image.

Automatic segmentation tools are available and roughly split into
two groups, supervised and unsupervised methods. The methods vary
and include: automatic gray level thresholding \cite{Kanade11}, the
watershed algorithm \cite{Vincent91} and Active Contours \cite{Bamford98,Meijering12tracking}.
Another approach is to support the segmentation algorithm with temporal
information from tracking algorithms as was proposed by \cite{Schiegg14,Arbelle15}.
All these methods assume some structure in the data that may not fit
every case.

Supervised methods, on the other hand, do not assume any structure
rather aim to learn it from the data. Classic machine learning methods generally
require two independent steps, feature extraction and classification.
In most cases the feature extraction is based either on prior knowledge
of the image properties such as in \cite{Su13} or general image properties
such as smoothing filters, edge filters, etc. 
A widely used toolbox which takes a pixel classification approach is Ilastik \cite{Sommer2011},
using a random forest classifier trained on predefined features extracted
from a user's scribbles on the image.

Recent developments in the computer vision community have shown the
strength Convolutional Neural Networks (CNNs) which surpass state of the art methods in object classification \cite{krizhevsky12}, 
semantic segmentation \cite{long2015} and many
other tasks. Recent attempts at cell segmentation using CNNs include
\cite{Ronneberger15,kraus2016}. The common ground of all CNN methods is the need for an extensive training set alongside a predefined loss function such as the cross-entropy (CE).

In this work we present a novel approach for microscopy cell segmentation inspired by the GAN \cite{goodfellow2014} and extension thereof \cite{radford2015unsupervised,mirza2014conditional,im2016generating,isola2016image}.
The GAN framework is based on two networks, a generator and a discriminator, trained simultaneously, with opposing objectives. This allows the discriminator to act as an abstract loss function in contrast to the common CE and $L_{1}$ losses. We propose a pair of adversarial networks, an \textbf{estimator} and a discriminator for the task of microscopy cell segmentation. Unlike the original GAN \cite{goodfellow2014}, we do not generate images from random noise vectors, rather estimate the underlaying variables of an image. The estimator learns to output some segmentation of the image while the discriminator learns to distinguish between expert manual segmentations
and estimated segmentations given the associated image. The discriminator is trained to \emph{minimize} a classification loss on two classes, manual and estimated, i.e. minimizing the similarity between the two. The estimator, on the other hand, is trained to \emph{maximize }the
discriminator's loss and effectively, maximize the similarity. 
In \cite{luc2016semantic}, semantic segmentation of natural images are generated for a set of predefined class. However, the main difference lays in our need to separate instances of a single class (cells) and not to separate different classes. The method also differs in choice of discriminator architecture and training method. 

Our contribution is three-fold. We expand the concept of the GAN
for the task of cell segmentation and in that reduce the dependency
on a selection of loss function. We propose a novel architecture for
the discriminator, referred to as the ``Rib Cage'' architecture
(See section \ref{subsec:Rib-Cage}), which is adapted
to the problem. The ``Rib Cage'' architecture includes several cross
connections between the image and the segmentation, allowing the network
to model complicated correlation between the two. Furthermore we show
that accurate segmentations can be achieved with a low number of
training examples therefore dramatically reducing the manual workload.

The rest of the paper is organized as follows. Section~\ref{sec:Proposed-Algorithm} defines the problem and elaborates on the proposed solution. Section~\ref{sec:Experiments} presents the results for both a common adversarial and non-adversarial loss compared to the proposed method,  showing promising
initial results. Section~\ref{sec:Summary} summarizes and concludes the work thus far.
\section{Methods}
\label{sec:Proposed-Algorithm}

\subsection{Problem Formulation}

\label{subsec:Problem-Formulation}

Let $\Omega$ define the image domain and let the image $I:\Omega\rightarrow\mathbb{R}^{+}$
be an example generated by the random variable $\mathcal{I}$. Our
objective is to partition the image into individual cells, where the
main difficulty is separating adjacent cells. Let the segmentation
image $\Gamma:\Omega\rightarrow\left\{ 0,1,2\right\} $ be a partitioning
of $\Omega$ to three disjoint sets, background, foreground (cell
nuclei) and cell contour, also generated by some random variable $\mathcal{S}$.
The two random variables are statistically dependent with some unknown
joint probability $P_{\mathcal{I},\mathcal{S}}$. The problem we address
can be formulated as the most likely partitioning $\hat{\Gamma}$
from the data $I$ given only a small number, $N$, of example pairs
$\left\{ I_{n},\Gamma_{n}\right\} _{n=1}^{N}$. Had $P_{\mathcal{S}|\mathcal{I}}\left(\Gamma|I\right)$
been known, the optimal estimator would be the Maximum Likelihood
(ML) estimator:

\begin{equation}
\hat{\Gamma}_{opt}=\arg\max_{\Gamma}P_{\mathcal{S}|\mathcal{I}}\left(\Gamma|I\right)\label{eq:Gamma-opt}
\end{equation}
However, since $P_{\mathcal{S}|\mathcal{I}}\left(\Gamma|I\right)$
is unknown and $\hat{\Gamma}_{opt}$ cannot be calculated, we learn
the near-optimal estimator of $\Gamma$ using the manual segmentation,
$\Gamma_{M}$, as our target.

\subsection{Estimation Network}

\label{subsec:Estimator}

We propose an estimator $\hat{\Gamma}=\mathcal{E}\left(I,\hat{\theta_{\mathcal{E}}}\right)$
in the form of a CNN with parameters $\theta_{\mathcal{E}}$. We wish
to train the estimator $\mathcal{E}$ such that the estimated $\hat{\Gamma}$
will be as close as possible to the optimal ML estimation $\hat{\Gamma}_{opt}$.
This is achieved by optimizing for some loss function $L_{\mathcal{E}}$ (defined in section \ref{subsec:Adversarial-Nets}): 
\begin{equation}
\hat{\theta}_{\mathcal{E}}=\arg\min_{\theta_{\mathcal{E}}}L_{\mathcal{E}}\left(\mathcal{E}\left(I,\theta_{\mathcal{E}}\right),\hat{\Gamma}_{opt}\right)\label{eq:Theta_E}
\end{equation}

\subsection{Adversarial Networks}

\label{subsec:Adversarial-Nets}

Unlike the GAN, aiming to \emph{generate examples} from an unknown
distribution, we aim to \emph{estimate the variables} of an unknown
conditional distribution $P_{\mathcal{S}|\mathcal{I}}\left(\Gamma|I\right)$.
Defining the loss $L_{\mathcal{E}}$ either in a supervised pixel-based
way, e.g. $L_{2}$ norm, or in an unsupervised global method, by  a cost functional that constrains partition into homogenous regions while minimizing the length of their boundaries, is usually not well defined. We
define the loss $L_{\mathcal{E}}$ by pairing our estimator with a
discriminator. Let $\mathcal{E}_{\theta_{\mathcal{E}}}$ and $\mathcal{D}_{\theta_{\mathcal{D}}}$
denote the estimator and discriminator respectively, both implemented
as CNN with parameters $\theta_{\mathcal{E}}$ and $\theta_{\mathcal{D}}$
respectively. The estimator aims to find the best estimation $\hat{\Gamma}$
of the partitioning $\Gamma$ given the image $I$. The discriminator
on the other hand tries to distinguish between $\Gamma_{M}$ and $\hat{\Gamma}$
given pairs of either $\left(I,\Gamma_{M}\right)$ or $\left(I,\hat{\Gamma}\right)$
and outputs the probability that the input is manual rather than estimated
denoted as $D\left(I,\hat{\Gamma}\right)$. As is in the GAN case,
the objectives of the estimator and the discriminator are exactly
opposing and so are the losses for training $\mathcal{E}_{\theta_{\mathcal{E}}}$
and $\mathcal{D}_{\theta_{\mathcal{D}}}$. We train $\mathcal{D}_{\theta_{\mathcal{D}}}$
to \emph{maximize} the probability of assigning the correct label
to both manual examples and examples estimated by $\mathcal{E}_{\theta_{\mathcal{E}}}$.
We simultaneously train $\mathcal{E}_{\theta_{\mathcal{E}}}$ to \emph{minimize}
the same probability, essentially trying to make $\hat{\Gamma}$ and
$\Gamma_{M}$ as similar as possible: 
\begin{equation}
L_{\mathcal{D}}=\mathbb{E}\left[\log\left(D\left(I,\Gamma_{M}\right)\right)\!+\!\log\left(1\!-\!D\left(I,\mathcal{E}_{\theta_{\mathcal{E}}}\left(I\right)\right)\right)\right]\label{eq:L_D}
\end{equation}

\begin{equation}
L_{\mathcal{E}}=\mathbb{E}\left[\log\left(D\left(I,\mathcal{E}_{\theta_{\mathcal{E}}}\left(I\right)\right)\right)\right]\label{eq:L_E}
\end{equation}
In other words, $\mathcal{E}_{\theta_{\mathcal{E}}}$ and $\mathcal{D}_{\theta_{D}}$
are players in a min-max game with the value function:

\begin{equation}
\min_{\mathcal{E}_{\theta_{\mathcal{E}}}}\max_{\mathcal{D}_{\theta_{D}}}\mathbb{E}\left[\log\left(D\left(I,\Gamma_{M}\right)\right)+\log\left(1-D\left(I,\mathcal{E}_{\theta_{\mathcal{E}}}\left(I\right)\right)\right)\right]\label{eq:min-max}
\end{equation}
The equilibrium is achieved when $\hat{\Gamma}$ and $\Gamma_{M}$
are similar such that the discriminator can not distinguish between
the pairs $\left(I,\hat{\Gamma}\right)$ and $\left(I,\Gamma_{M}\right)$.

\subsection{Implementation Details}

\label{subsec:Implementation Details}

\subsubsection{Estimator Network Architecture}

\label{subsec:Network-Architecture}

The estimator $\mathcal{E}_{\theta_{\mathcal{E}}}$ net is designed
as a five layer fully CNN, each layer is constructed of a convolution
followed by batch normalization and leaky-ReLU activation. The output
of the estimator is an image with the same size as the input image
with three channels corresponding to the probability that a pixel
belongs to the background, foreground or cell contour.

\subsubsection{Discriminator Network Architecture}

\label{subsec:Rib-Cage}

\begin{figure}

\noindent \begin{centering}
\includegraphics[trim={20bp 190bp 10bp 50bp},clip,width=0.99\columnwidth]{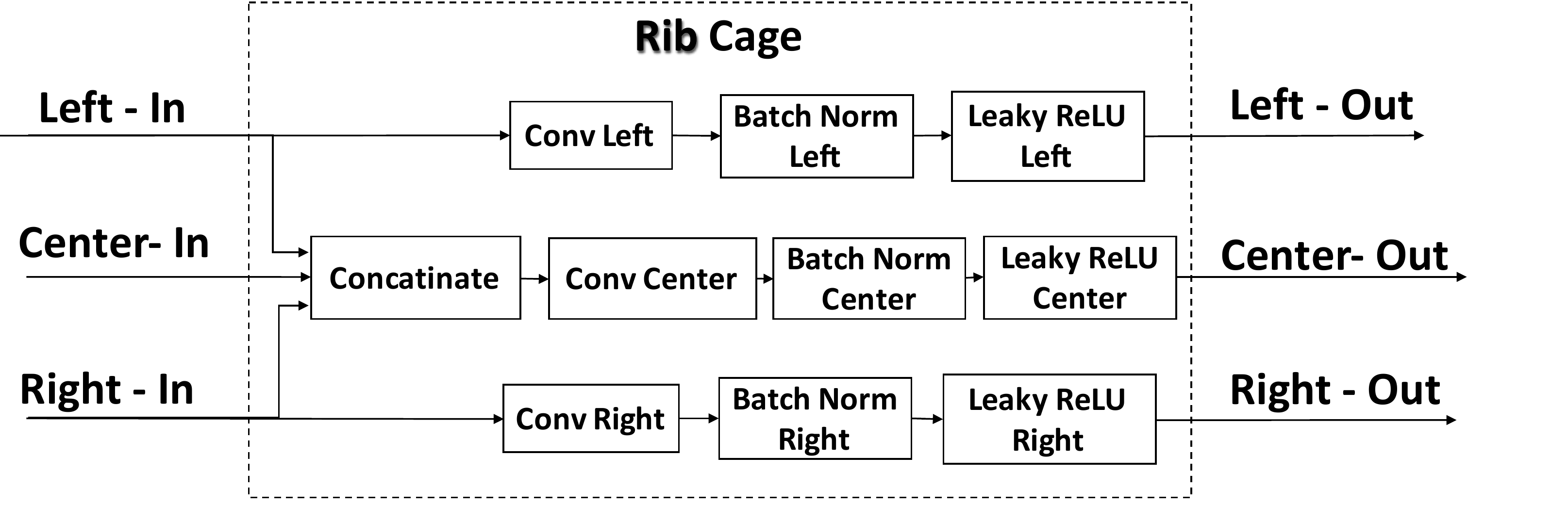} 
\par\end{centering}
\protect\caption{The design of the basic building block for the discriminator. Each
block has three inputs and three outputs. \label{fig:Rib-Cage}}
\end{figure}

\begin{figure}
\noindent \begin{centering}
\includegraphics[width=0.9\columnwidth]{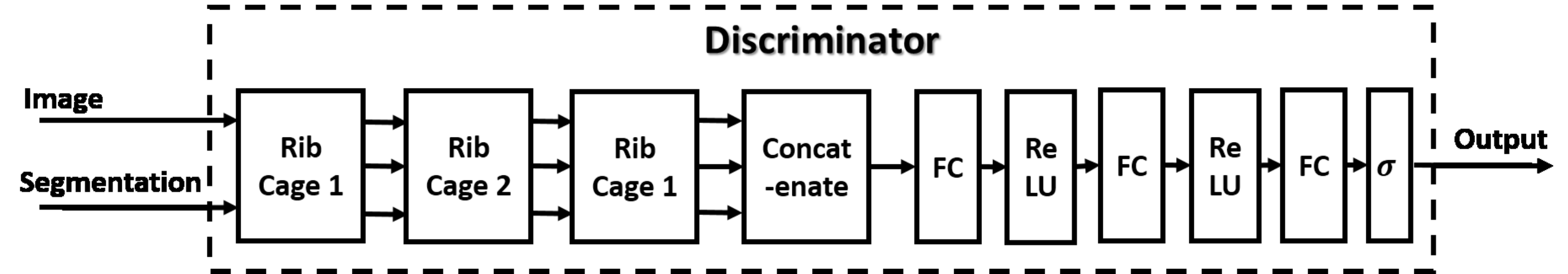} 
\par\end{centering}
\protect\caption{The design of the Discriminator $\mathcal{D}_{\theta_{\mathcal{D}}}$.
Three ``Rib Cage'' blocks (see Figure \ref{fig:Rib-Cage})
are followed by two FC layers with ReLU activations and a last FC
layer with a sigmoid activation, $\sigma$. The Center-In channel of the first Rib Cage block is omitted. \label{fig:Disc}}
\end{figure}

The discriminator $\mathcal{D}_{\theta_{D}}$ is designed with a more
complex structure. The discriminators task is to distinguish manual and
estimated segmentation images given a specific gray level (GL) image.
The question arrises of how to design the discriminator architecture which can get both the GL and segmentation images as input. A basic design s that of a classification CNN where both images are concatenated in the channel axis, as done in \cite{isola2016image}. However, we believe that this approach is not optimal for our needs since, for this task, the discriminator should be able match high level features from the GL and segmentation images. Yet these features may have very
different appearances. For example, an edge of a cell in the GL image appear as a transition from white to black while the same edge in the segmentation image appears as a thin blue line. This difference requires the network to learn individual filters for each semantic region. Then, finding correlations between the two is a more feasible task. 
For these reasons we designed a specific architecture, referred to as a ``Rib Cage'' architecture, which has three channels. The first and second
channels get inputs from the GL channel and segmentation channel respectively, each channel calculates feature maps using a convolutional layer, we refer to these channels as the ``Ribs''. The third
channel, referred to  as the ``Spine'',  gets a concatenation of inputs from both the GL and segmentation
channels and matches feature maps (i.e correlations). See Figure \ref{fig:Rib-Cage} for an illustration of the ``Rib Cage'' block. The discriminator
is designed as three consecutive ``Rib Cage'' blocks followed
by two fully-connected (FC) layers with leaky-ReLU activations and
a final FC layer with one output and a sigmoid activation for classification.
Figure \ref{fig:Disc} illustrates the discriminator design. The architecture
parameters for the convolution layers are describes as $C\left(kernel~size,\#\,filters\right)$
and FC layers as $F\left(\#\,filters\right)$. The parameters for
the estimator: $C\left(9,16\right),$ $C\left(7,32\right),$ $C\left(5,64\right),$
$C\left(4,64\right),$ $C\left(1,3\right)$. The discriminator spine used half the number of filters as the ribs: $C\left(9,8\right),$
$C\left(5,32\right),$ $C\left(3,64\right),$ $C\left(4,64\right),$
$F\left(64\right),$ $F\left(64\right),$ $F\left(1\right)$. 

\subsubsection{Data}

We trained the networks on the H1299 data set \cite{cohen08} consisting
of $72$ frames of size $512\times640$ pixels. Each frame captures approximately
50 cells. Manual annotation of $15$ randomly selected frames was
done by an expert. The annotated set was split into a training set
and validation set. The training set was subsampled to $N_{Train}\in[1,2,4,11]$
examples for training which were augmented using randomly
cropped areas of size $64\times64$ pixels along with random
flip and random rotation. The images were annotated using three labels
for the background (red), cell nucleus (green) and nucleus contour (blue) encoded as RGB images.
\section{Experiments and Results}

\label{sec:Experiments}

\begin{figure}
\begin{center}
\setlength{\tabcolsep}{0mm}
\renewcommand{\arraystretch}{0.3}
\begin{subfigure}[b]{.03\linewidth}
\begin{tabular}{cc}
\rotatebox{90}{~~~~~~Full Image}\\
\rotatebox{90}{~~~Zoom}\\
~
\end{tabular}
\end{subfigure}
\begin{subfigure}[b]{.31\linewidth}
\begin{tabular}{cc}
\includegraphics[trim={50bp 40bp 0bp 102bp},clip, width=1\columnwidth]{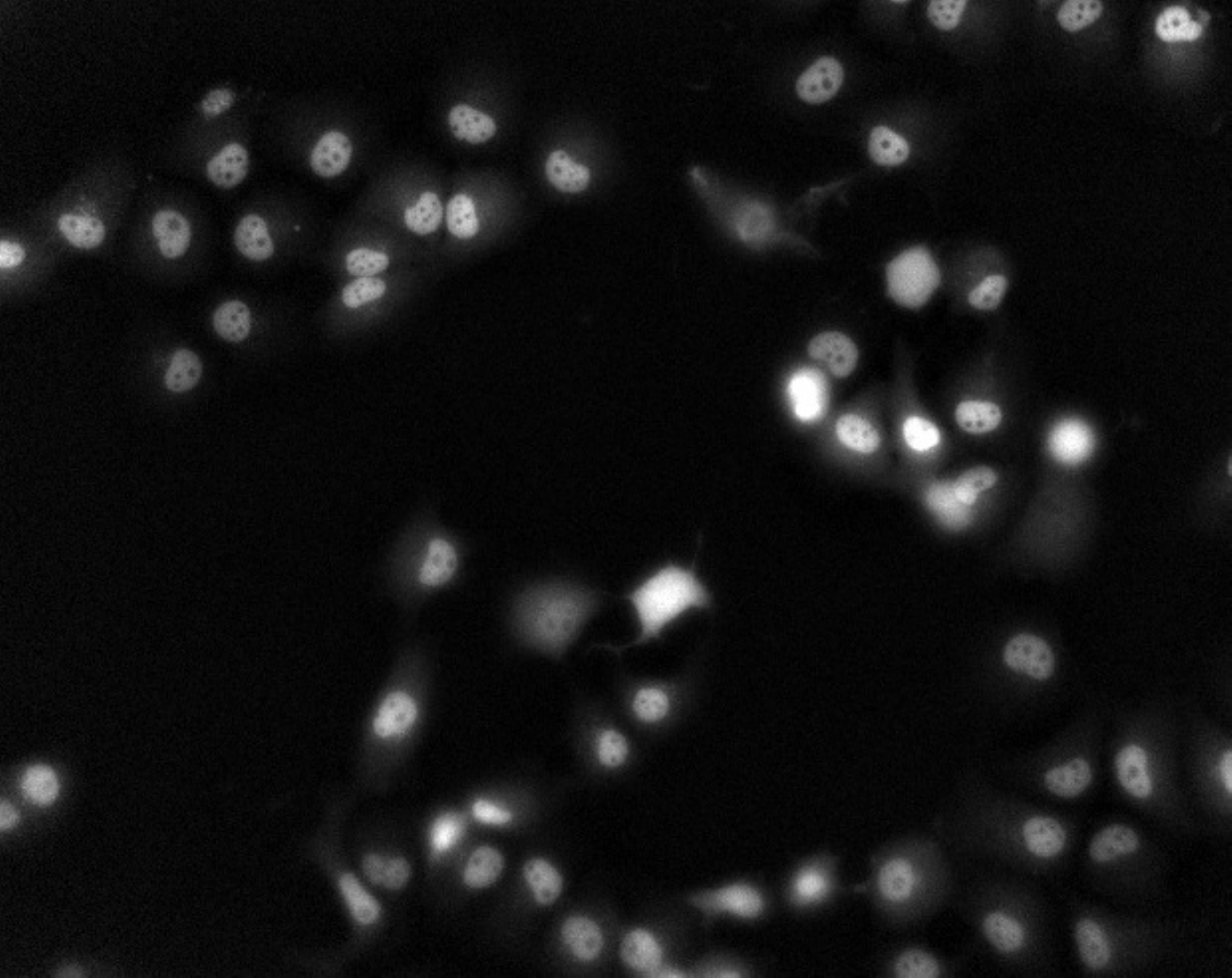}   \\
\includegraphics[trim={200bp 57bp 340bp 397bp},clip,width=1\columnwidth]{Alon_Lab_H1299_t_60_y_1_x_1}  \\
Image
\end{tabular}
\end{subfigure}
\begin{subfigure}[b]{.31\linewidth}
\begin{tabular}{c}
\includegraphics[trim={50bp 40bp 0bp 102bp},clip, width=1\columnwidth]{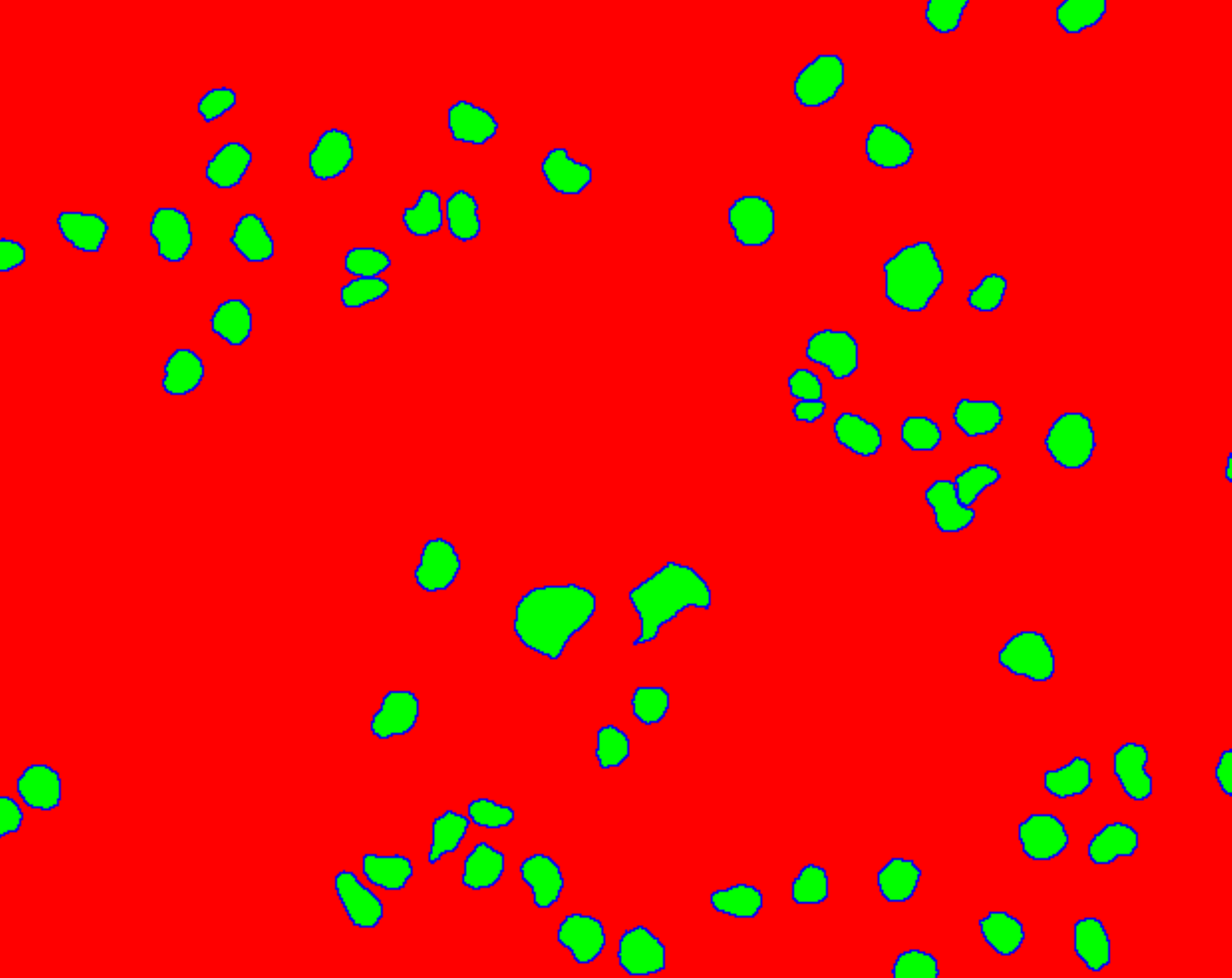}     \\
\includegraphics[trim={200bp 57bp 340bp 397bp},clip,width=1\columnwidth]{GT_Alon_Lab_H1299_t_60_y_1_x_1}\\
Manual Seg
\end{tabular}
\end{subfigure}
\begin{subfigure}[b]{.31\linewidth}
\begin{tabular}{c}
\includegraphics[trim={50bp 40bp 0bp 102bp},clip, width=1\columnwidth]{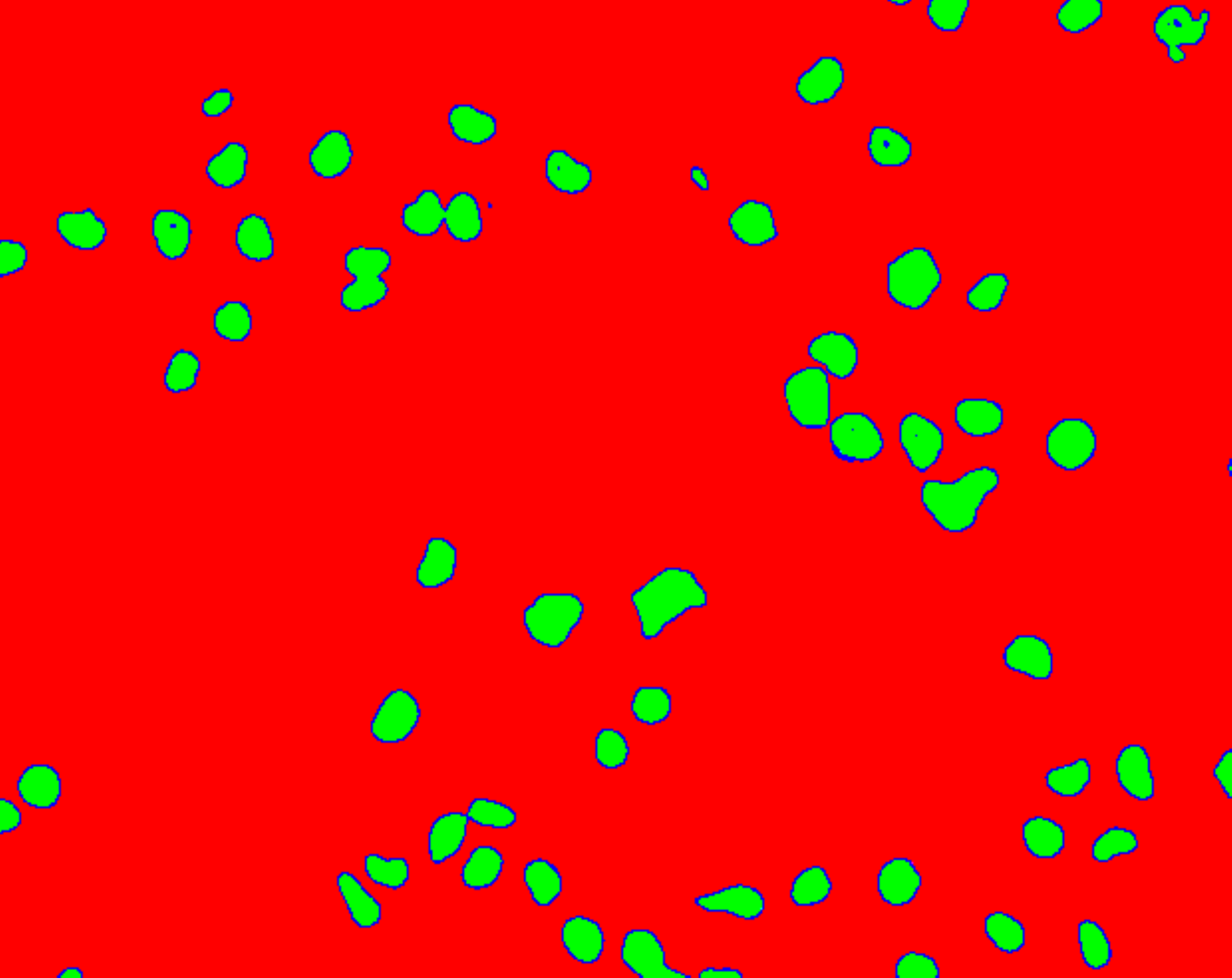}   \\
\includegraphics[trim={200bp 57bp 340bp 397bp},clip,width=1\columnwidth]{ex_1_Alon_Lab_H1299_t_60_y_1_x_1}   \\
$N_{Train}=1$
\end{tabular}
\end{subfigure}

\begin{subfigure}[b]{.03\linewidth}
\begin{tabular}{cc}
\rotatebox[origin=c]{90}{~~~~~~Full Image}\\
\rotatebox[origin=c]{90}{~~~Zoom}\\
~
\end{tabular}
\end{subfigure}
\begin{subfigure}[b]{.31\linewidth}
\begin{tabular}{c}
\includegraphics[trim={60bp 40bp 0bp 102bp},clip, width=1\columnwidth]{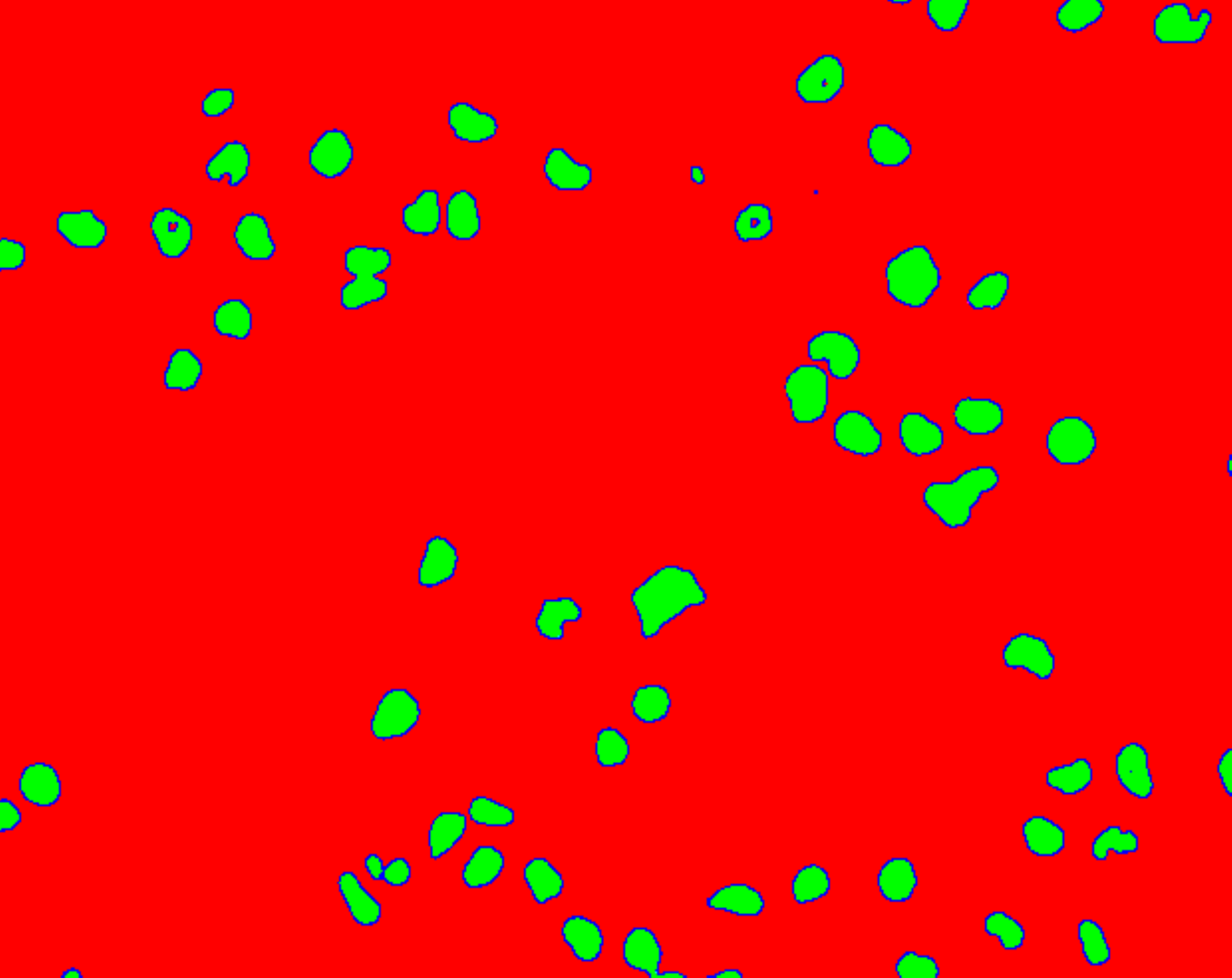}     \\
\includegraphics[trim={200bp 57bp 340bp 397bp},clip,width=1\columnwidth]{ex_2_Alon_Lab_H1299_t_60_y_1_x_1}\\
$N_{Train}=2$
\end{tabular}
\end{subfigure}
\begin{subfigure}[b]{.31\linewidth}
\begin{tabular}{c}
\includegraphics[trim={50bp 40bp 0bp 102bp}, clip, width=1\columnwidth]{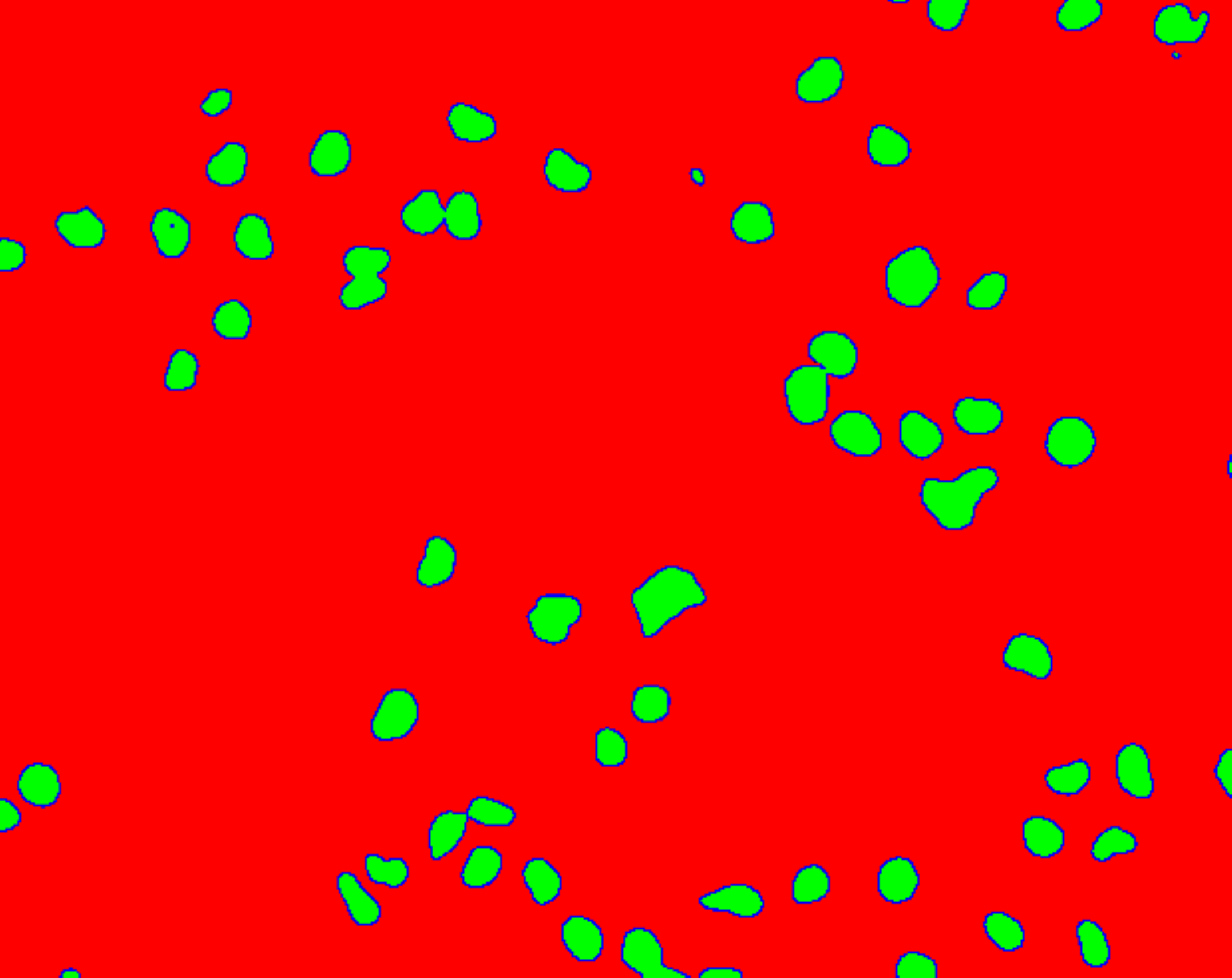}   \\
\includegraphics[trim={200bp 57bp 340bp 397bp},clip,width=1\columnwidth]{ex_4_Alon_Lab_H1299_t_60_y_1_x_1}\\
$N_{Train}=4$
\end{tabular}
\end{subfigure}
\begin{subfigure}[b]{.31\linewidth}
\begin{tabular}{c}
\includegraphics[trim={50bp 40bp 0bp 102bp},clip, width=1\columnwidth]{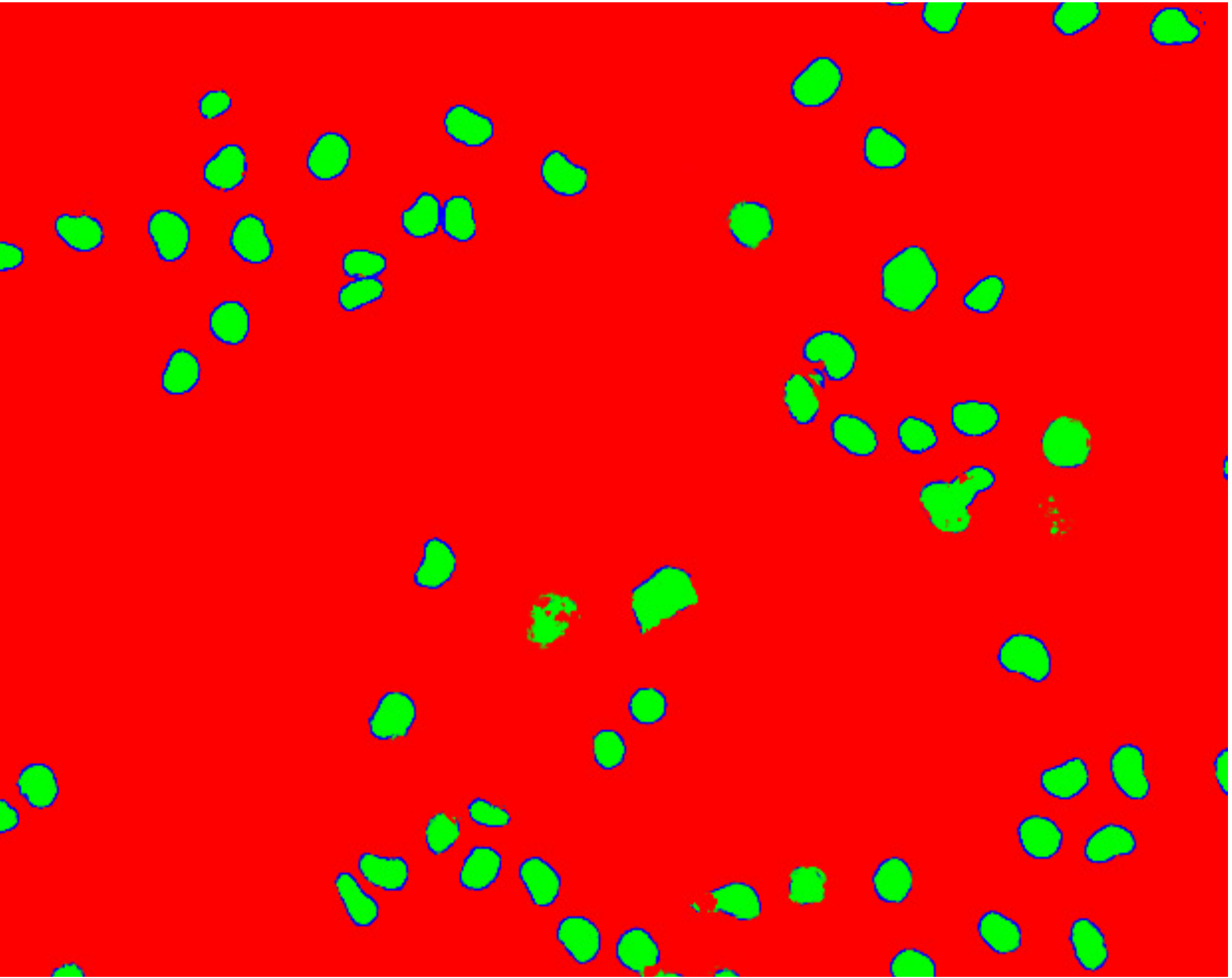}   \\
\includegraphics[trim={200bp 57bp 340bp 397bp},clip,width=1\columnwidth]{ex_no_adv_Alon_Lab_H1299_t_60_y_1_x_1} \\
CE Loss  
\end{tabular}
\end{subfigure}
\caption{\label{fig:Results}Segmentation example of a validation
image given a different number of training examples. The odd and even rows
show the full image and a zoomed area respectively.
Notice that in all cases the cells in the second row were correctly separated even
though they are very close together. The bottom right shows the
result when training with the CE loss.}
\end{center}

\end{figure}

\begin{table*}
\begin{centering}
\begin{tabular}{|c||>{\centering}p{0.09\columnwidth}|>{\centering}p{0.1\columnwidth}|>{\centering}p{0.1\columnwidth}|>{\centering}p{0.1\columnwidth}||>{\centering}p{0.115\columnwidth}|>{\centering}p{0.115\columnwidth}|>{\centering}p{0.115\columnwidth}|}
\hline 
\hspace{-2cm}  & \textbf{ADV-1}  & \textbf{ADV-2}  & \textbf{ADV-4}  & \textbf{ADV-11}  & \textbf{CE - 11}  & \textbf{Class Disc} & \textbf{Ilastik}\tabularnewline
\hline 
\hline 
Prec  & \textbf{89.9\%}  & 85.4\%  & 86.8\%  & 85.8\%  & 83.6\%  & 78.7\% & 81.2\%\tabularnewline
\hline 
Rec  & \textbf{82\%}  & \textbf{87.2\%}  & \textbf{86.8\%}  & \textbf{86.5\%}  & 86.4\%  & 81.14\% & 80.2\%\tabularnewline
\hline 
F  & \textbf{85.8\%}  & \textbf{86.3\%}  & \textbf{86.8\%}  & \textbf{86.1\%}  & 84.8\%  & 79.9\% & 80.7\%\tabularnewline
\hline 
J  & \textbf{80.6\%}  & \textbf{75.8\%}  & \textbf{77.4\%}  & \textbf{74.6\%}  & 72.1\%  & 60.2\% & 68.4\%\tabularnewline
\hline 
\end{tabular}\\
 ~ 
\par\end{centering}
\centering{}\caption{\label{tab:Results}\textbf{Quantitative Results:} \small{Each column represents
an experiment with a different number of training examples, ADV-$N_{Train}$. CE Loss-11 and ClassDisc are experiments using the same estimator network trained with the pixel-based CE loss and a simple classification discriminator respectively. The last column is the comparison
to the state of the art tool, Ilastik \cite{Sommer2011}. The rows
are the results for individual cell segmentation. As is explained
in \cite{Schiegg14} True positives (TP) are cells with Jaccard measure
greater than 0.5. False positives (FP) are automatic segmentation
not appearing in the manual segmentation and false negatives (FN)
is the opposite. The measures are defines as $Prec=\frac{TP}{TP+FP}$,
$Rec=\frac{TP}{TP+FN}$, $F-Measure=2\frac{Prec*Rec}{Prec+Rec}$. J indicates the mean Jaccard measure for individual cells.}}
\end{table*}

We conducted four experiments, training the networks with different
values for $N_{Train}$. All other parameters were set identically.
We evaluated the segmentation using the as described in the caption of Table~\ref{tab:Results}.
We compared the adversarial training regime to the common CE loss,
training only the estimator. We furthermore evaluate our choice of RibCage discriminator versus a classification architecture (VGG16 \cite{simonyan2014}).
We also compared our results to state of the art segmentation tool,
Ilastik \cite{Sommer2011}. The  manual annotation were done by an expert. The quantitative results of the
individual cell segmentation are detailed in Table~\ref{tab:Results}.
Note that the amount of images in the training data had little effect
on the results. 
Figure~\ref{fig:Results} shows an example
of a segmented frame. It is clear that the networks learned a few
distinct properties of the segmentation. First, each cell
is encircled by a thin blue line. Second, the shape of the contour follows
the true shape of the cell. Some drawbacks are still seen where
two cells completely touch and the boundary is difficult to determine.
\section{Summary}

\label{sec:Summary}
In this work we propose a new concept for microscopy cell segmentation
using CNN with adversarial loss. The contribution of such an approach
is two-fold. First, the loss function is automatically defined
as it is learned along side the estimator, making this a simple to
use algorithm with no tuning necessary. Second, we show that this
method is robust to low number of training examples surpassing.

The quantitative results, as well as the visual results, show clearly
that both the estimator and our unique ``Rib Cage'' discriminator
learn both global and local properties of the segmentation, i.e the
shape of the cell and the contour surrounding the cell, and the fitting
of segmentation edges to cell edges. These properties could not be
learned using only a pixel-bases CE loss as is commonly done. 

{\small{}{  \bibliographystyle{IEEEbib}
\bibliography{DeepSeg}
 } }{\small \par}
\end{document}